%% file: paper.tex
\documentclass[british,conference,twocolumn]{IEEEtran}
\usepackage[T1]{fontenc}
\usepackage[latin9]{inputenc}
\usepackage{array}
\usepackage{verbatim}
\usepackage{rotating}
\usepackage{multirow}
\usepackage{amsmath}
\usepackage{amssymb}
\usepackage{graphicx}

\makeatletter

\providecommand{\tabularnewline}{\\}

\makeatletter
\def\markboth#1#2{\def\leftmark{\@IEEEcompsoconly{\sffamily}\MakeUppercase{\protect#1}}%
\def\rightmark{\@IEEEcompsoconly{\sffamily}\MakeUppercase{\protect#2}}}
\makeatother

\usepackage[british]{babel}

\usepackage{acronym}
\usepackage{graphicx}
\usepackage{cite}       

\usepackage{algorithm}
\usepackage{algorithmicx}
\usepackage{algpseudocode}

\usepackage{booktabs}
\usepackage{multirow}

\graphicspath{{Figures/}}
\usepackage{pgfplots}
\pgfplotsset{compat=newest}
\pgfplotsset{plot coordinates/math parser=false}
\usetikzlibrary{plotmarks}

\usepackage{exscale,relsize}

\input{operatornames.tex}

\input{svgsupport.tex}
\input{acronyms.tex}


\newlength\figureheight
\newlength\figurewidth

\makeatother

\usepackage{babel}
\begin{document}

\title{Design and Implementation of an Inertial Navigation System for Pedestrians
Based on a Low-Cost MEMS IMU}

\author{\IEEEauthorblockN{Francesco Montorsi\IEEEauthorrefmark{1}, Fabrizio
Pancaldi\IEEEauthorrefmark{2} and Giorgio M. Vitetta\IEEEauthorrefmark{1}}
\IEEEauthorblockA{\IEEEauthorrefmark{1}Department of Engineering
``Enzo Ferrari'', University of Modena and Reggio Emilia\\
 Modena, Italy, Email: \{francesco.montorsi, giorgio.vitetta\}@unimore.it}
\IEEEauthorblockA{\IEEEauthorrefmark{2}Department of Science
and Methods for Engineering, University of Modena and Reggio Emilia\\
Modena, Italy, Email: fabrizio.pancaldi@unimore.it}}
\maketitle
\begin{abstract}
Inertial navigation systems for pedestrians are infrastructure-less
and can achieve sub-meter accuracy in the short/medium period. However,
when low-cost inertial measurement units (IMU) are employed for their
implementation, they suffer from a slowly growing drift between the
true pedestrian position and the corresponding estimated position.
In this paper we illustrate a novel solution to mitigate such a drift
by: a) using only accelerometer and gyroscope measurements (no magnetometers
required); b) including the sensor error model parameters in the state
vector of an extended Kalman filter; c) adopting a novel soft heuristic
for foot stance detection and for zero-velocity updates. Experimental
results evidence that our inertial-only navigation system can achieve
similar or better performance with respect to pedestrian dead-reckoning
systems presented in related studies, although the adopted IMU is
less accurate than more expensive counterparts.
\end{abstract}

\section{Introduction\label{sec:intro}}

In an \ac{INS} the position of a mobile agent is tracked by means
of an \ac{IMU} carried by the agent itself. \ac{IMU}-based \acp{INS}
can provide a \emph{low-cost} and \emph{infrastructure-less} solution
to accurate indoor navigation in the short/medium term. Unluckily,
in the medium/long term they usually suffer from a ``drift'' phenomenon
\cite{Jim2010a}, which originates from the noise and from any small
bias in the accelerations and angular velocities sensed by the \ac{IMU}.

Recently, substantial attention has been devoted to \ac{PDR} \acp{INS},
where the prior knowledge of human walking patterns is exploited to
reset, at least partially, the accumulated errors due to various error
sources (e.g., the time-variant biases of \acp{IMU}). This approach
has been first proposed in \cite{Foxlin2005}, where the periods during
which the pedestrian's foot is still on the ground are detected and
exploited to introduce some corrections (the so-called \acp{ZUPT})
in the tracking filter. Further advances have been developed in \cite{Jim2010a,Seco2011,Ramon2012,Krach2008b}.
In particular, in \cite{Jim2010a} and \cite{Seco2011} an \ac{EKF}
processing \ac{IMU} measurements exploits various heuristics to
compensate for the drift due to time-variant biases and measurement
noise. In \cite{Ramon2012} and \cite{Krach2008b}, instead, additional
measurements (from RFID devices) are adopted to mitigate the drift
phenomenon.

In this manuscript, starting from the methods and the results illustrated
in \cite{Jim2010a,Hol2008}, we develop a novel \ac{INS} based
only on a low-cost \ac{IMU} which performs \ac{PDR} employing
an \ac{EKF}. Unlike previous approaches, the proposed solution
relies on:
\begin{enumerate}
\item Accelerometer and gyroscope measurements only (magnetometer sensors
are often completely unreliable in indoor environments and other technologies
for accurate localization are expensive).
\item A rigorous approach to the kinematic modelling of \ac{IMU} measurements;
this involves the use of a large EKF state vector, including both
physical variables (e.g., agent position and heading) and quantities
referring to the \acp{SEM}.
\item A new soft (rather than hard) heuristic for foot stance detection
which increases the overall accuracy of the INS.
\end{enumerate}
This manuscript is organized as follows. In Section \ref{sec:experimental_setup},
the employed \ac{IMU} and its calibration procedure are described.
The proposed PDR-INS is illustrated in Section \ref{sec:map_unaware_ins},
whereas its performance is assessed in Section \ref{sec:indoor_nav_tests}.
Finally, in Section \ref{sec:conc} some conclusions are provided.

\emph{Notations}: The \emph{probability density function} (pdf) of
a \emph{random vector} (rv) $\mathbf{R}$ evaluated at the point $\mathbf{r}$
is denoted as $f(\mathbf{r})$; $\mathcal{N}\left(\mathbf{r};\mathbf{m},\boldsymbol{\Sigma}\right)$
denotes the pdf of a Gaussian rv $\mathbf{R}$ having mean $\mathbf{m}$
and covariance matrix $\boldsymbol{\Sigma}$, evaluated at the point
$\mathbf{r}$; $\left\Vert \mathbf{x}\right\Vert $ denotes the $L^{2}$
norm of vector $\mathbf{x}$; the expressions $\{\mathbf{x}_{i}\}_{i=1}^{k}$
and $\mathbf{x}_{1:k}$ both denote the sequence $\mathbf{x}_{1},\mathbf{x}_{2},...,\mathbf{x}_{k}$.
$g\triangleq9.80665\mpssq$ denotes the gravitational acceleration;
finally, $\odot$ denotes the quaternion multiplication \cite{Kuipers1999}.

\section{IMU Description and Calibration\label{sec:experimental_setup}}

In our PDR INS a mobile agent is equipped with a low-cost \ac{IMU},
called RazorIMU \cite{Razor_IMU_documents} and fixed on one of his/her
feet using shoes' laces (e.g., see \cite{Jim2010a,Seco2011,Ramon2012,Krach2008b}).
It is important to note that the IMU-sensed quantities are expressed
in \emph{body }(or \emph{sensor})\emph{ frame}, i.e., they are referred
to a right-handed coordinate frame centered on the IMU with axes parallel
to the sensor sides; this frame is different from the so called \emph{navigation
frame}, which is a right-handed coordinate frame centered on some
point of the navigation map and whose $x$ and $y$ axes are parallel
to Earth ground and $z$ axis points away from Earth.

The RazorIMU is a programmable device equipped with 3-axis accelerometers,
gyroscopes and magnetometers; our firmware outputs their measurements
in ``raw mode'', i.e., as integer numbers, so that a calibration
procedure is required. The tri-axial accelerometer calibration procedure
we adopted is similar to that described in \cite{Batista2011a,Pa2012},
but does not require any additional hardware (besides the IMU itself).
It relies on the \ac{SEM} (assuming a still sensor) \cite{Batista2011a,Pa2012}
\begin{equation}
\mathbf{a}^{m}=\mathbf{G}_{a}\mathbf{a}+\mathbf{b}^{a}+\mathbf{n}^{a}\label{eq:acc_sem}
\end{equation}
where $\mathbf{a}^{m}\in\mathbb{Z}^{3}$ is the vector of \emph{measured}
accelerations (in body frame), $\mathbf{a}\in\mathbb{R}^{3}$ is the
vector of \emph{true} accelerations (in body frame), $\mathbf{G}_{a}\in\mathbb{R}^{3\times3}$
is the \emph{gain matrix} (diagonal if only scale factors are accounted
for, or a generic invertible matrix if cross-couplings are also accounted
for), $\mathbf{b}^{a}\in\mathbb{R}^{3}$ is the \emph{bias} vector,
in body frame, and $\mathbf{n}^{a}\in\mathbb{R}^{3}$ is the \emph{noise}
vector (in body frame) and is assumed to be \ac{AGN} with covariance
matrix $\boldsymbol{\Sigma}_{a}=\sigma_{a}^{2}\mathbf{I}_{3}$. The
calibration task for the accelerometer consists of estimating $\mathbf{G}_{a}$
and $\mathbf{b}^{a}$ on the basis of $N\cdot P$ measured vectors
$\{\{\mathbf{a}_{i,p}^{m}\}_{i=1}^{N}\}_{p=1}^{P}$ (in our setup
$N=500$ and $P=16$), referring to $P$ unknown orientations of the
still sensor (in the navigation frame). The optimal (in the \textit{mean
square error} sense) estimators $\hat{\mathbf{G}}_{a}$ and $\hat{\mathbf{b}}^{a}$
of the terms $\mathbf{G}_{a}$ and $\mathbf{b}^{a}$ appearing in
(\ref{eq:acc_sem}) are given by
\begin{equation}
\left(\hat{\mathbf{G}}_{a},\hat{\mathbf{b}}^{a}\right)=\arg\min_{\left(\tilde{\mathbf{G}},\tilde{\mathbf{b}}\right)\in\mathcal{G}\times\mathcal{B}}\sum_{p=1}^{P}r_{p}\left(\tilde{\mathbf{G}},\tilde{\mathbf{b}},\bar{\mathbf{a}}_{p}^{m}\right)\label{eq:acc_calibration_estimator}
\end{equation}
where $\bar{\mathbf{a}}_{p}^{m}$ is the mean of the measurements
$\left\{ \mathbf{a}_{i,p}^{m}\right\} _{i=1}^{N}$ and 
\begin{equation}
r_{p}\left(\tilde{\mathbf{G}},\tilde{\mathbf{b}},\bar{\mathbf{a}}_{p}^{m}\right)\triangleq\min_{\tilde{\mathbf{a}}_{p}\in\mathcal{A}}\left\Vert \tilde{\mathbf{G}}\tilde{\mathbf{a}}_{p}+\tilde{\mathbf{b}}-\bar{\mathbf{a}}_{p}^{m}\right\Vert ^{2}\label{eq:acc_calibration_residual}
\end{equation}

Once calibration is completed, the true acceleration vector may be
estimated as
\begin{equation}
\hat{\mathbf{a}}=\hat{\mathbf{G}}_{a}^{-1}\left(\mathbf{a}^{m}-\hat{\mathbf{b}}^{a}\right).\label{eq:acc_calibrated_output}
\end{equation}

Regarding the gyroscopes, a \ac{SEM} similar to (\ref{eq:acc_sem}),
i.e., 
\begin{equation}
\boldsymbol{\omega}^{m}=\mathbf{G}_{\omega}\boldsymbol{\omega}+\mathbf{b}^{\omega}+\mathbf{n}^{\omega}\label{eq:gyro_sem}
\end{equation}
has been exploited to devise our calibration procedure; here $\boldsymbol{\omega}^{m}\in\mathbb{Z}^{3}$
is the vector of measured angular velocities (in body frame), $\boldsymbol{\omega}\in\mathbb{R}^{3}$
is the vector of true angular velocities (in body frame), $\mathbf{G}_{\omega}\in\mathbb{R}^{3\times3}$
is the gain matrix (diagonal if only scale factors are accounted for,
or a generic invertible matrix if cross-couplings are also accounted
for), $\mathbf{b}^{\omega}\in\mathbb{R}^{3}$ is the bias vector (in
body frame), and $\mathbf{n}^{\omega}\in\mathbb{R}^{3}$ is the noise
vector (in body frame) and is assumed to be \ac{AGN} with covariance
matrix $\boldsymbol{\Sigma}_{\omega}=\sigma_{\omega}^{2}\mathbf{I}_{3}$.
Similarly to (\ref{eq:acc_calibrated_output}), the true vector $\boldsymbol{\omega}$
is estimated as 
\begin{equation}
\hat{\boldsymbol{\omega}}=\hat{\mathbf{G}}_{\omega}^{-1}\left(\boldsymbol{\omega}^{m}-\hat{\mathbf{b}}^{\omega}\right)\label{eq:gyro_calibrated_output}
\end{equation}
where $\hat{\mathbf{G}}_{\omega}$and $\hat{\mathbf{b}}^{\omega}$
denote the estimated bias vector and gain matrix of the gyroscope,
respectively. However, unlike accelerometer calibration, calibration
of gyroscopes requires an expensive dedicated hardware platform, so
that $\hat{\mathbf{b}}^{\omega}=0$ and the value provided in the
gyroscope datasheet \cite{Razor_IMU_documents} for $\hat{\mathbf{G}}_{\omega}$
have been adopted.

\section{The PDR INS\label{sec:map_unaware_ins}}

Our \ac{INS} performs similarly to some other navigation systems
described in the technical literature (e.g., see \cite{Jim2010a}),
but is based on a different approach and, in particular, on a set
of rigorous kinematic equations relating the quantities sensed by
the IMU with its orientation and 3D position. After describing the
structure of the state vector, the dynamic models and the measurement
models, we describe the use of an \ac{EKF} for estimating the posterior
distribution of the state vector. Finally, we focus on a soft algorithm
for foot stance detection.

\subsection{State Vector\label{sub:ekf_state_vector}}

In our INS the state vector $\mathbf{x}_{k}$ of the mobile agent
wearing the IMU is defined as 
\begin{equation}
\mathbf{x}_{k}\triangleq\left[\mathbf{p}_{k},\mathbf{v}_{k},\mathbf{a}_{k},q_{k}^{\nb},\mathbf{a}_{k}^{b},\boldsymbol{\omega}_{k}^{\bbn},\mathbf{b}_{k}^{a},\mathbf{b}_{k}^{\omega}\right]^{T}\in\mathbb{R}^{D}\label{eq:ekf_state_vector}
\end{equation}
where $k$ is the time-index of the discrete-time tracking filter
for navigation, $\mathbf{p}_{k}\in\mathbb{R}^{3}$, $\mathbf{v}_{k}\in\mathbb{R}^{3}$
and $\mathbf{a}_{k}\in\mathbb{R}^{3}$ are the position, the velocity
and the acceleration of the IMU sensor, measured in $\meter$, $\mps$
and $\mpssq$, respectively; $q_{k}^{\nb}\in\mathbb{H}_{1}$ is a
(random) \emph{quaternion} representing the transformation which produces,
given a vector in navigation coordinates, a vector in body coordinates
\cite{Kuipers1999}; $\mathbf{a}_{k}^{b}$ and $\boldsymbol{\omega}_{k}^{\bbn}\in\mathbb{R}^{3}$
are the acceleration (in body frame) and the angular velocity (from
body to navigation frame, resolved in body coordinate frame \cite{Hol2008}),
measured in $\mpssq$ and $\radps$, respectively; $\mathbf{b}_{k}^{a}\in\mathbb{R}^{3}$
and $\mathbf{b}_{k}^{\omega}\in\mathbb{R}^{3}$ are the bias vectors
of the accelerometer and of the gyroscope, expressed in body coordinate
frame and measured in $\mpssq$ and $\radps$, respectively; finally,
$D=25$ is the size of $\mathbf{x}_{k}$. 

Note that: a) $\mathbf{a}_{k}^{b}$ and $\boldsymbol{\omega}_{k}^{\bbn}$
are the (noisy) observable variables; b) all the other variables can
be \emph{pseudo-observed} to enhance the system stability whenever
the foot is (approximately) still; c) $\mathbf{b}_{k}^{a}$ and $\mathbf{b}_{k}^{\omega}$
represent ``fine'' bias vectors, and play a complementary role with
respect to $\hat{\mathbf{b}}^{a}$ and $\hat{\mathbf{b}}^{\omega}$,
respectively, which account for time-variant and turn-on dependent
biases; d) the vector $\mathbf{x}_{k}$ (\ref{eq:ekf_state_vector})
has an heterogeneous structure, since it consists of quantities of
interest for the end-user of the INS (namely, $\mathbf{p}_{k}$ and
$\mathbf{v}_{k}$), quantities relating $\mathbf{p}_{k}$ and $\mathbf{v}_{k}$
to the sensor outputs (namely, $\mathbf{a}_{k}$, $q_{k}^{\nb}$,
$\mathbf{a}_{k}^{b}$, $\boldsymbol{\omega}_{k}^{\bbn}$) and quantities
related to the SEMs of accelerometers and gyroscopes ($\mathbf{b}_{k}^{a}$
and $\mathbf{b}_{k}^{\omega}$, respectively); e) including the IMU-sensed
quantities ($\mathbf{a}_{k}^{b}$, $\boldsymbol{\omega}_{k}^{\bbn}$)
in $\mathbf{x}_{k}$ is important since impulsive noise (due to hardware
instability of low-cost sensors) may affect the IMU output and an
accurate dynamic modelling of $(k+1)$-th step sensor orientation
($q_{k+1}^{\nb}$) requires the knowledge of the $k$-th step angular
velocity ($\boldsymbol{\omega}_{k}^{\bbn}$).

\subsection{Dynamic and Measurement Models\label{sub:ekf_models}}

The dynamic models adopted for the elements of $\mathbf{x}_{k}$ (\ref{eq:ekf_state_vector})
can be summarised as follows. The Taylor-expansion models (e.g., see
\cite[Sec. 4.3]{Hol2008,RongLi2003})
\begin{equation}
\mathbf{p}_{k+1}=\mathbf{p}_{k}+\mathbf{v}_{k}\cdot T_{s}+\frac{1}{2}\mathbf{a}_{k}\cdot T_{s}^{2}+\mathbf{n}_{p,k}\label{eq:pos_dyn_model}
\end{equation}
and
\begin{equation}
\mathbf{v}_{k+1}=\mathbf{v}_{k}+\mathbf{a}_{k}\cdot T_{s}+\mathbf{n}_{v,k}\label{eq:vel_dyn_model}
\end{equation}
have been employed for the vectors $\mathbf{p}_{k}$ and $\mathbf{v}_{k}$,
respectively; here $T_{s}$ denotes the sampling period of the INS
($1/100\hz$ in our case) and the vectors $\mathbf{n}_{p,k}$ and
$\mathbf{n}_{v,k}$ are \ac{AGN} terms affecting $\mathbf{p}_{k}$
and $\mathbf{v}_{k}$, respectively. The model 
\begin{equation}
\mathbf{a}_{k+1}=R^{T}\left(q_{k}^{\nb}\right)\mathbf{a}_{k}^{b}+\mathbf{g}+\mathbf{n}_{a,k}\label{eq:acc_dyn_model}
\end{equation}
has been used for $\mathbf{a}_{k}$, where $R\left(q_{k}^{\nb}\right)$
is the rotation matrix associated with the quaternion $q_{k}^{\nb}$
(and thus representing the transformation from navigation to body
frame), $\mathbf{g}\triangleq\left[0,0,-g\right]^{T}$ is the gravity
vector in the navigation frame and $\mathbf{n}_{a,k}$ is AGN.

A further model relates the orientation of the sensor (represented
by $q_{k}^{\nb}$) to the angular velocity $\boldsymbol{\omega}_{k}^{\bbn}$
and is given by (e.g., see \cite[Eq. (4.11), Eq. (4.20d)]{Hol2008}
and \cite[Sec. 11.5]{Kuipers1999}) 
\begin{equation}
q_{k+1}^{\nb}=\exp\left(-\frac{T}{2}\boldsymbol{\omega}_{k}^{\bbn}\right)\odot q_{k}^{\nb}+\mathbf{n}_{q,k}\label{eq:quat_dyn_model}
\end{equation}
where $\mathbf{n}_{q,k}$ is AGN.

Finally, the simple ``random walk'' models
\begin{align}
\mathbf{a}_{k+1}^{b} & =\mathbf{a}_{k}^{b}+\mathbf{n}_{a,k}\label{eq:bodyacc_dyn_model}\\
\boldsymbol{\omega}_{k+1}^{\bbn} & =\boldsymbol{\omega}_{k}^{\bbn}+\mathbf{n}_{\omega,k}\label{eq:bodygyro_dyn_model}\\
\mathbf{b}_{k+1}^{a} & =\mathbf{b}_{k}^{a}+\mathbf{n}_{b^{a},k}\label{eq:accbias_dyn_model}\\
\mathbf{b}_{k+1}^{\omega} & =\mathbf{b}_{k}^{\omega}+\mathbf{n}_{b^{\omega},k}\label{eq:gyrobias_dyn_model}
\end{align}
have been selected for the filtered states $\mathbf{a}_{k}^{b}$ and
$\boldsymbol{\omega}_{k}^{\bbn}$, and for the sensor biases $\mathbf{b}_{k}^{a}$
and $\mathbf{b}_{k}^{\omega}$, where the $\left\{ \mathbf{n}_{\cdot,k}\right\} $
terms denote AGN contributions. 

Regarding measurements models, simple linear relations involving only
quantities in the body frame may be adopted, thanks to the structure
chosen for $\mathbf{x}_{k}$ (\ref{eq:ekf_state_vector}): 
\begin{equation}
\mathbf{z}_{k}^{f}=\mathbf{a}_{k}^{b}+\mathbf{b}_{k}^{a}+\mathbf{m}_{a,k}\label{eq:meas_model_acc}
\end{equation}
\begin{equation}
\mathbf{z}_{k}^{\omega}=\boldsymbol{\omega}_{k}^{\bbn}+\mathbf{b}_{k}^{\omega}+\mathbf{m}_{\omega,k}\label{eq:meas_model_angvel}
\end{equation}
Here $\mathbf{z}_{k}^{f}=\hat{\mathbf{a}}$ (see (\ref{eq:acc_calibrated_output}))
and $\mathbf{z}_{k}^{\omega}=\hat{\boldsymbol{\omega}}$ (see (\ref{eq:gyro_calibrated_output}))
denote the \emph{calibrated} force and angular velocity measurements
provided by the IMU and the vectors $\mathbf{m}_{\mathit{a,k}}$,
$\mathbf{m}_{\mathit{\omega,k}}$ represent the AGN terms affecting
the measurements. 

The \emph{dynamic }models (\ref{eq:pos_dyn_model})-(\ref{eq:gyrobias_dyn_model})
can be summarised as
\begin{align}
f(\mathbf{x}_{k+1}|\mathbf{x}_{k}) & =\mathcal{N}\left(\mathbf{x}_{k+1};\mathbf{q}\left(\mathbf{x}_{k}\right),\mathbf{Q}\right)\label{eq:dyn_model}
\end{align}
whereas the \emph{measurement }models (\ref{eq:meas_model_acc})-(\ref{eq:meas_model_angvel})
can be summarised as\emph{ }
\begin{equation}
f(\mathbf{z}_{k}|\mathbf{x}_{k})=\mathcal{N}\left(\mathbf{z}_{k};\mathbf{r}(\mathbf{x}_{k}),\mathbf{R}\right)\label{eq:meas_model}
\end{equation}
where $\mathbf{z}_{k}\triangleq[\mathbf{z}_{k}^{f},\mathbf{z}_{k}^{\omega}]^{T}\in\mathbb{R}^{M}$
(with $M=6$), the vector functions $\mathbf{q}\left(\cdot\right)$
and $\mathbf{r}(\cdot)$ are defined by (\ref{eq:pos_dyn_model})-(\ref{eq:gyrobias_dyn_model})
and by (\ref{eq:meas_model_acc})-(\ref{eq:meas_model_angvel}), respectively,
and $\mathbf{Q}$ and $\mathbf{R}$ are $D\times D$ and $M\times M$
diagonal covariance matrices for the \ac{AGN} terms. Regarding
these matrices, it is worth mentioning that a) they may have a strong
impact on the EKF stability and b) the choice of their diagonal values
can be based, in practice, on some careful tuning procedure (involving
$D+M=31$ parameters).

\subsection{The EKF\label{sub:ekf}}

The goal of the INS is the sequential estimation of the hidden state
vector $\mathbf{x}_{k}$ representing the mobile agent given the sequence
of IMU measurements $\left\{ \mathbf{z}_{0:k}\right\} $, i.e., the
sequential estimation of the posterior pdf $f\left(\mathbf{x}_{k}|\mathbf{z}_{0:k}\right)$.
Since our dynamic model is non-linear (see (\ref{eq:acc_dyn_model})
and (\ref{eq:quat_dyn_model})), a non-linear filter, such as an \ac{EKF},
needs to be employed to solve this problem. It is important to mention
that: a) the EKF alternates a \emph{prediction step} with an \emph{update
step}; b) it estimates the first two moments of the posterior pdf
$f(\mathbf{x}_{k}|\mathbf{z}_{0:k})$, namely, the mean state vector
$\hat{\mathbf{x}}_{k}^{\ekf}$ and the state vector covariance matrix
$\hat{\mathbf{P}}_{k}^{\ekf}$, in a recursive fashion. In particular,
given $\hat{\mathbf{x}}_{k}^{\ekf}$ and $\hat{\mathbf{P}}_{k}^{\ekf}$,
the EKF estimates (\emph{prediction} step) \cite{Kalman1960}
\begin{align*}
\hat{\mathbf{x}}_{k+1|k}^{\ekf} & =\mathbf{q}\left(\hat{\mathbf{x}}_{k}^{\ekf}\right)\qquad\hat{\mathbf{P}}_{k+1|k}^{\ekf}=\mathbf{J}_{k}^{q}\hat{\mathbf{P}}_{k}^{\ekf}\left(\mathbf{J}_{k}^{q}\right)^{T}+\mathbf{Q}
\end{align*}
where $\hat{\mathbf{x}}_{k+1|k}^{\ekf}$ and $\hat{\mathbf{P}}_{k+1|k}^{\ekf}$
denote the $(k+1)$-th state mean and covariance, respectively, which
can be predicted on the basis of the information available at the
$k$-th step; here $\mathbf{J}_{k}^{q}\triangleq\left.\frac{\partial\mathbf{q}(\mathbf{x})}{\partial\mathbf{x}}\right|_{\hat{\mathbf{x}}_{k}}$
is the $D\times D$ Jacobian matrix%
\footnote{This matrix cannot be put in a simple analytical form, so that its
evaluation requires use of computer algebra systems.%
} for our (non-linear) dynamic model. Then, the EKF evaluates\emph{
}(\emph{update} step) \cite{Kalman1960}:
\begin{align*}
\mathbf{s}_{k+1|k} & =\mathbf{z}_{k}-\mathbf{r}\left(\hat{\mathbf{x}}_{k+1|k}^{\ekf}\right)\\
\mathbf{S}_{k+1|k} & =\mathbf{J}_{k}^{r}\hat{\mathbf{P}}_{k+1|k}^{\ekf}\left(\mathbf{J}_{k}^{r}\right)^{T}+\mathbf{R}\\
\mathbf{K}_{k+1|k} & =\hat{\mathbf{P}}_{k+1|k}^{\ekf}\left(\mathbf{J}_{k}^{r}\right)^{T}\mathbf{S}_{k+1|k}^{-1}\\
\hat{\mathbf{x}}_{k+1}^{\ekf} & =\hat{\mathbf{x}}_{k+1|k}^{\ekf}+\mathbf{K}_{k+1|k}\mathbf{s}_{k+1|k}\\
\hat{\mathbf{P}}_{k+1}^{\ekf} & =(\mathbf{I}-\mathbf{K}_{k+1|k}\mathbf{J}_{k}^{r})\hat{\mathbf{P}}_{k+1|k}^{\ekf}
\end{align*}
where $\mathbf{r}_{k+1|k}$ is the innovation residual and $\mathbf{S}_{k+1|k}$
its estimated covariance matrix, $\mathbf{K}_{k+1|k}$ is the Kalman
gain, $\hat{\mathbf{x}}_{k+1}^{\ekf}$ is the new estimate of the
state vector mean and $\hat{\mathbf{P}}_{k+1}^{\ekf}$ is its estimated
covariance matrix; moreover, $\mathbf{J}_{k}^{r}\triangleq\left.\frac{\partial\mathbf{r}\left(\mathbf{x}\right)}{\partial\mathbf{x}}\right|_{\hat{\mathbf{x}}_{k+1|k}}$
is the $M\times D$ Jacobian matrix for the measurement model.

It is worth noting that: a) in any EKF, at the end of the $k$-th
iteration only the quantities $\hat{\mathbf{x}}_{k}^{\ekf}$ and $\hat{\mathbf{P}}_{k}^{\ekf}$
need to be saved and this substantially simplifies the INS implementation;
b) given these quantities, the posterior distribution $f(\mathbf{x}_{k}|\mathbf{z}_{0:k})$
is estimated by the EKF as $\mathcal{N}\left(\mathbf{x}_{k};\hat{\mathbf{x}}_{k}^{\ekf},\hat{\mathbf{P}}_{k}^{\ekf}\right)$;
c) the initialisation of the INS represent a critical task, since
initial errors cannot be mitigated by the EKF. As far as the last
point is concerned, $\hat{\mathbf{P}}_{0}^{\ekf}=\mathbf{Q}$ has
been selected for the initial covariance matrix, whereas the initial
state vector $\hat{\mathbf{x}}_{0}^{\ekf}$ has been estimated assuming
the foot still in a known position; unfortunately, further details
cannot be provided for space limitations.

\subsection{Foot Stance Detection\label{sub:ekf_heuristics}}

Even if the EKF illustrated in the previous Paragraph includes the
sensor biases $\mathbf{b}_{k}^{a}$ and $\mathbf{b}_{k}^{\omega}$
in $\mathbf{x}_{k}$, due to the lack of robust models and, in particular,
to the lack of bias observations, the tracking of such quantities
mitigates but does not completely compensate for sensor inaccuracies.
In practice, the residual biases may quickly disrupt the INS tracking
since their effects accumulate over time. The effects of these error
sources can be mitigated exploiting some a priori knowledge about
the typical human walking pattern and, in particular, the fact at
the end of each step the foot lies approximately still on the ground
for a short period (typically, $0.1-0.2\second$); during such a period,
the value of most of the elements of $\mathbf{x}_{k}$ are known a
priori and the EKF state can be adjusted accordingly. In practice,
the EKF can be provided with some ``pseudo-measurements'', usually
known as \acp{ZUPT} \cite{Foxlin2005}, whenever a detection algorithm,
processing the IMU measurements in parallel to the EKF, detects a
``foot still event''. In our work, a foot stance detection algorithm
inspired by \cite[Sec. II.C]{Jim2010a} has been used. This algorithm
evaluates four logical ``condition signals'' $\left\{ C_{i}^{1},C_{i}^{2},C_{i}^{3},C_{i}^{4}\right\} $
associated with the IMU measurements $\mathbf{z}_{k}$ and generated
as
\begin{align*}
C_{i}^{1}\triangleq & \begin{cases}
1 & \gamma_{a,\mymin}<\left\Vert \mathbf{z}_{i}^{f}\right\Vert <\gamma_{a,\mymax}\\
0 & \text{otherwise}
\end{cases}\\
C_{i}^{2}\triangleq & \begin{cases}
1 & \sigma\left(\mathbf{z}_{i-S:i+S}^{f}\right)<\sigma_{a,\mymax}\\
0 & \text{otherwise}
\end{cases}\\
C_{i}^{3}\triangleq & \begin{cases}
1 & \left\Vert \mathbf{z}_{i}^{\omega}\right\Vert <\gamma_{\omega,\mymax}\\
0 & \text{otherwise}
\end{cases}\\
C_{i}^{4}\triangleq & \begin{cases}
1 & \sigma\left(\mathbf{z}_{i-S:i+S}^{\omega}\right)<\sigma_{\omega,\mymax}\\
0 & \text{otherwise}
\end{cases}
\end{align*}
for $i\in\{k-F,...,k+F\}$, where $\sigma(\mathbf{x}_{1},\mathbf{x}_{2},...,\mathbf{x}_{N})$
denotes the standard deviation of the magnitude of the vectors $\left\{ \mathbf{x}_{1},\mathbf{x}_{2},...,\mathbf{x}_{N}\right\} $,
$F$ is the size of the windows used for step detection, $S$ is the
size of the window used for the computation of $\sigma(\cdot)$, and
$\gamma_{a,\mymax}$, $\sigma_{a,\mymax}$ , $\gamma_{\omega,\mymax}$
and $\sigma_{\omega,\mymax}$ represent proper thresholds. An \emph{hard}
detection algorithm based on the condition signals defined above has
been proposed in\cite[Sec. II.C]{Jim2010a}; it decides that the foot
is ``still'', during the $k$-th time step, if $\sum_{j=k-F}^{k+F}C_{j}^{1}C_{j}^{2}C_{j}^{3}>\frac{F}{2}$.
Here, we propose to use a \emph{soft }variant whose output is the
\ac{SFS} signal 
\[
\textrm{SFS}_{k}\triangleq\frac{1}{F}\sum_{i=k-F}^{k+F}C_{i}^{1}C_{i}^{2}C_{i}^{3}C_{i}^{4}
\]
which ranges, for the $k$-th time step, from zero (moving foot) to
one (the foot is very likely to be still on the ground). Then, whenever
$\textrm{SFS}_{k}>\gamma_{\sfs}$ ($\gamma_{\sfs}\in[0;1]$ is a fixed
threshold), a ``foot still event'' begins and the EKF is fed with
the pseudo-measurements
\begin{align}
\left[\hat{\mathbf{p}}_{s}^{\ekf}\right]_{1:2}=\mathbf{z}^{xy} & =\left[\mathbf{p}_{k}\right]_{1:2}+\mathbf{m}_{xy,k}\label{eq:pseudo_xy}\\
0=z^{z} & =\left[\mathbf{p}_{k}\right]_{3}+m_{z,k}\label{eq:pseudo_z}\\
\mathbf{0}=\mathbf{z}^{v} & =\mathbf{v}_{k}+\mathbf{m}_{v,k}\label{eq:pseudo_vel}\\
\mathbf{0}=\mathbf{z}^{a} & =\mathbf{a}_{k}+\mathbf{m}_{a,k}\label{eq:pseudo_acc}\\
-\mathbf{g}=\mathbf{z}^{a^{b}} & =R^{T}\left(q_{k}^{\nb}\right)\mathbf{a}_{k}^{b}+\mathbf{m}_{a^{b},k}\label{eq:pseudo_quat_and_body_acc}\\
\left\Vert \mathbf{g}\right\Vert =\mathbf{z}^{a^{b}} & =\left\Vert \mathbf{a}_{k}^{b}\right\Vert +m_{a^{b},k}\label{eq:pseudo_body_acc_norm}
\end{align}
\begin{align}
\mathbf{0}=\mathbf{z}^{\omega} & =\boldsymbol{\omega}_{k}^{\bbn}+\mathbf{m}_{\omega,k}\label{eq:pseudo_angvel}\\
\hat{\mathbf{a}}=\mathbf{z}_{k}^{b^{a}} & =\mathbf{b}_{k}^{a}-R\left(q_{k}^{\nb}\right)\mathbf{g}+\mathbf{m}_{b^{a},k}\label{eq:pseudo_body_acc_bias}\\
\hat{\boldsymbol{\omega}}=\mathbf{z}_{k}^{b^{\omega}} & =\mathbf{b}_{k}^{\omega}+\mathbf{m}_{b^{\omega},k}\label{eq:pseudo_angvel_bias}
\end{align}
where $s$ is the time step corresponding to the beginning of the
current ``foot still event'' (so that $\hat{\mathbf{p}}_{s}^{\ekf}$
represents the position where the foot is still) and the vectors $\left\{ \mathbf{m}_{\cdot,k}\right\} $
denote the AGN terms of the pseudo-measurement models. To avoid discontinuities
in the tracked path, the variance of these noise terms is modulated
in a \emph{soft} way on the basis of the $\textrm{SFS}_{k}$ signal.
In practice, the soft ZUPT pseudo-measurement model 
\begin{equation}
f\left(\mathbf{z}_{k}^{p}|\mathbf{x}_{k}\right)=\mathcal{N}\left(\mathbf{z}_{k}^{p};\mathbf{r}^{p}(\mathbf{x}_{k}),\left[1+K^{p}(1-\textrm{SFS}_{k})\right]\mathbf{R}^{p}\right)\label{eq:ekf_zupt_model}
\end{equation}
is adopted where $\mathbf{z}_{k}^{p}\triangleq[\mathbf{z}^{xy},z^{z},\mathbf{z}^{v},...,\mathbf{z}_{k}^{b^{\omega}}]^{T}\in\mathbb{R}^{M^{p}}$
(with $M^{p}=20$); $\mathbf{R}^{p}$ is the diagonal covariance matrix
collecting all variance values of the $\left\{ \mathbf{m}_{\cdot,k}\right\} $
terms; the vector function $\mathbf{r}^{p}(\cdot)$ can be easily
derived from (\ref{eq:pseudo_xy})-(\ref{eq:pseudo_angvel_bias});
$\mathbf{J}_{k}^{r^{p}}\triangleq\left.\frac{\partial\mathbf{r}^{p}\left(\mathbf{x}\right)}{\partial\mathbf{x}}\right|_{\hat{\mathbf{x}}_{k+1|k}}$
is the $M^{p}\times D$ Jacobian matrix associated with the measurement
model (\ref{eq:ekf_zupt_model}) and $K^{p}$ is a parameter introduced
to modulate the variance of ZUPT pseudo-measurements. Note that the
is the value of $K^{p}$, the higher will be the variances associated
to $\mathbf{z}_{k}^{p}$ pseudo-measurement when $\textrm{SFS}_{k}=\gamma_{\sfs}$;
then, as $\textrm{SFS}_{k}$ goes from $\gamma_{\sfs}$ to $1$, the
variances associated to $\mathbf{z}_{k}^{p}$ decrease smoothly to
the values collected in $\mathbf{R}^{p}$. This approach ensures that
the tracked state vector $\hat{\mathbf{x}}_{k}^{\ekf}$ smoothly transitions
to the ``reset'' values defined by (\ref{eq:pseudo_xy})-(\ref{eq:pseudo_angvel_bias}),
when ZUPTs are injected in the EKF.

Finally, it is worth noting that the pseudo-measurements (\ref{eq:pseudo_xy}),
(\ref{eq:pseudo_z}), (\ref{eq:pseudo_vel}), (\ref{eq:pseudo_body_acc_bias})
and (\ref{eq:pseudo_angvel_bias}) allow to ``observe'' otherwise
unobservable state vector components and thus to ``reset'' errors
they might contain.

\section{Indoor Navigation Tests\label{sec:indoor_nav_tests}}

An experimental campaign has been carried out to acquire various sets
of measurements generated by an agent equipped with the IMU described
in Section \ref{sec:experimental_setup} and repeating the same test
trajectory $N_{rep}=10$ times in an indoor environment. These measurement
sets have been stored on a notebook and then processed offline. The
test trajectory contains long straight lines, $90\deg$ turns, short
and long stops (e.g., to turn on/off lights, to open/close doors,
etc); the initial and final positions coincide and the \ac{TTD}
$L_{\ttd}$ associated to such test walk is $L_{\ttd}\simeq300\meter$.
\begin{figure}
\begin{centering}
\includegraphics[width=3.4in]{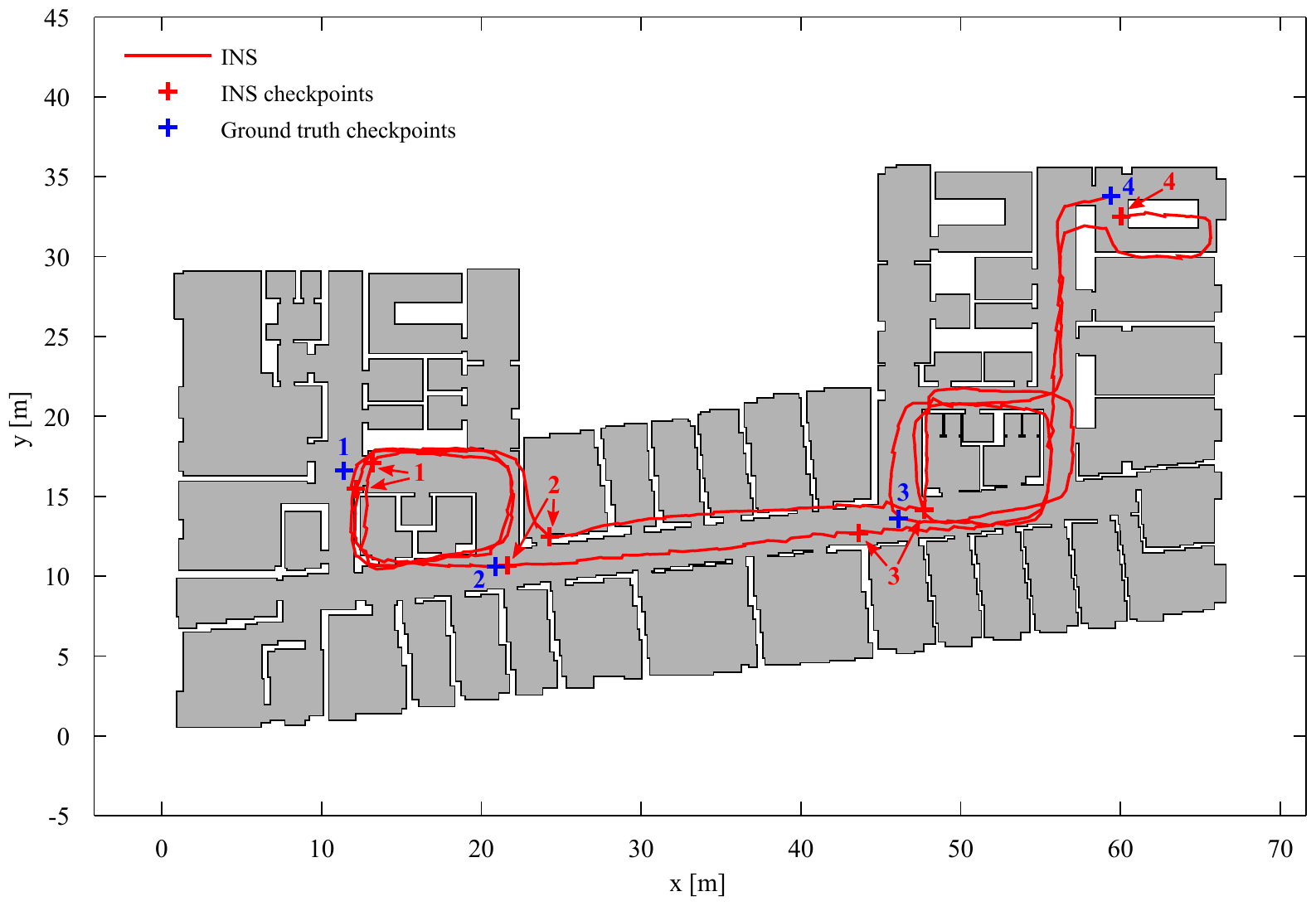}
\par\end{centering}

\centering{}%
\protect\caption{The agent trajectory estimated by the INS described in Section \ref{sec:map_unaware_ins}
for a specific repetition. The checkpoints approximating the ground
truth are indicated by blue crosses; the corresponding points estimated
by the INS are indicated by red crosses. \label{fig:path1_results}}
\end{figure}

An example of the resulting INS-estimated agent path%
\footnote{Note that the values of the parameters introduced throughout the paper
and used to produce such results have been\emph{ }properly \emph{tuned}
by means of an automated search procedure, but the resulting values
cannot be shown for space limitations; see \cite{Montorsi2013} for
more details.%
} is shown in Fig. \ref{fig:path1_results}. It is easy to recognize
that the ``drift'' phenomenon mentioned in Section \ref{sec:intro}
is present, although in a very limited amount (specially considering
the many loops walked in the same verse by the agent).

To quantify the performance of the INS usually the quantity $\epsilon_{\ttd}\triangleq\frac{1}{L_{\ttd}}\left\Vert \hat{\mathbf{p}}_{0}^{\ekf}-\hat{\mathbf{p}}_{N_{s}}^{\ekf}\right\Vert $
is exploited \cite{Jim2010a}, where $N_{s}$ is the last time step
in the recorded measurement set; of course, for a closed test path
and an ideal INS, $\epsilon_{\ttd}=0$. In our tests, for the $N_{rep}$
repetitions of the test walk, the $\epsilon_{\ttd}$ figure of merit
was $\{1.4,0.5,4.7,1.5,7.2,1.6,8.8,10.1,10.4,12.1\}$; these results
can be compared with those obtained in \cite[Table II]{Jim2010a};
in such contribution, when the magnetometer is not employed, the reported
range for $\epsilon_{\ttd}$ is $2-10$. These results show that our
INS achieves similar performance to that of \cite{Jim2010a}, despite
the key difference that in \cite{Jim2010a} the Xsens MTi IMU has
been employed. Such an IMU has higher accuracy (and higher costs)
than the RazorIMU; to quantify such a difference, the noise of the
IMU sensors can be modelled analysing, by means of the Allan variance
method, long sequences of sensor outputs acquired while the sensor
is still. In our case, 24h of RazorIMU accelerometer and gyroscope
data, acquired at the sampling frequency $f_{s}=100\hz$ have been
recorded and analysed; the results, in terms of the standard $N$
and $B$ coefficients representing \emph{acceleration/angular velocity
random walk }(ARW) and \ac{BI} noise contributions, are listed
in Table \ref{tab:gyro_comparison}, together with the results reported
in \cite[Table III]{Hoflinger2012} for the Xsens MTi IMU. The comparison
between the two IMUs shows that: a) the Xsens MTi has better matching
among the sensors mounted on the $x$, $y$ and $z$ axis; b) the
Xsens MTi IMU offers much better accelerometer performance. Moreover,
it is important to note that the RazorIMU calibration has been carried
out at a fixed temperature while the Xsens IMUs employ temperature-dependent
calibration factors. 
\begin{table}
\centering{}%
\begin{tabular}{c|c|c|c}
 & IMU & ARW/BI coefficient & Unit\tabularnewline
\hline 
\hline 
\multirow{4}{*}{\begin{turn}{90}
gyro
\end{turn}} & \multirow{2}{*}{RazorIMU} & $N=[5.2,12.1,5.6]\cdot10^{-3}$ & $\left(\degps\right)/\sqrt{\hz}$\tabularnewline
\cline{3-4} 
 &  & $B=[3,18,4.4]\cdot10^{-3}$ & $\degps$\tabularnewline
\cline{2-4} 
 & \multirow{2}{*}{Xsens MTi} & $N=[45,41,36]\cdot10^{-3}$ & $\left(\degps\right)/\sqrt{\hz}$\tabularnewline
\cline{3-4} 
 &  & $B=[7,7,5]\cdot10^{-3}$ & $\degps$\tabularnewline
\hline 
\multirow{4}{*}{\begin{turn}{90}
accel.
\end{turn}} & \multirow{2}{*}{RazorIMU} & $N=[5.5,5.1,7.6]\cdot10^{-3}$ & $\left(\mpssq\right)/\sqrt{\hz}$\tabularnewline
\cline{3-4} 
 &  & $B=[609,590,732]\cdot10^{-6}$ & $\mpssq$\tabularnewline
\cline{2-4} 
 & \multirow{2}{*}{Xsens MTi} & $N=[900,950,850]\cdot10^{-6}$ & $\left(\mpssq\right)/\sqrt{\hz}$\tabularnewline
\cline{3-4} 
 &  & $B=[230,270,290]\cdot10^{-6}$ & $\mpssq$\tabularnewline
\hline 
\end{tabular}\protect\caption{Comparison between the RazorIMU and the Xsens MTi IMU. \label{tab:gyro_comparison}}
\end{table}

In summary, the values of $\epsilon_{\ttd}$ characterizing our INS
are comparable to the values reported in \cite[Table II]{Jim2010a}
(when the magnetometer sensors are not used) although we employed
an IMU with worse noise and bias characteristics (of course, our IMU
is also cheaper and thus lowers system costs).

\section{Conclusions\label{sec:conc}}

In this manuscript, a novel INS has been derived integrating exact
kinematic models, SEM in the EKF state vector and a novel \emph{soft
}heuristic to detect foot steps. Our experimental tests have evidenced
that: a) a good accuracy can be achieved in tracking a mobile agent
on the short/medium period; b) our INS performs similarly to other
state-of-art INS PDR solutions but uses a lower-cost IMU and does
not employ magnetometers which are often unreliable in indoor environments.
Future work will focus the integration of map-awareness and radio
measurements in the proposed INS in order to further improve robustness
and long-term accuracy.

\bibliographystyle{IEEEtran}

\end{document}

%% file: operatornames.tex
\global\long\def\nb{\scriptscriptstyle{n \triangleright b}}

\global\long\def\bbn{\scriptscriptstyle{b, b \triangleright n}}

\newcommand{\textmu}[1]{\text{\small{#1}}}

\global\long\def\meter{\,\textmu{m}}

\global\long\def\hz{\,\textmu{Hz}}

\global\long\def\mps{\,\textmu{m}/\textmu{s}}
\global\long\def\mpssq{\,\textmu{m}/\textmu{s}^2}
\global\long\def\radps{\,\textmu{rad}/\textmu{s}}

\global\long\def\degps{\,{}^{\circ}/\textmu{s}}

\global\long\def\second{\,\textmu{s}}

\global\long\def\mymax{\text{\tiny{max}}}
\global\long\def\mymin{\text{\tiny{min}}}

\global\long\def\ekf{\text{\tiny{EKF}}}

\global\long\def\sfs{\text{\tiny{SFS}}}

\global\long\def\ttd{\text{\tiny{TTD}}}

%% file: svgsupport.tex
\newcommand{\executeiffilenewer}[3]{%
\ifnum\pdfstrcmp{\pdffilemoddate{#1}}%
{\pdffilemoddate{#2}}>0%
{\immediate\write18{#3}}\fi%
}

%% file: acronyms.tex
\newacro{GPS}{global positioning system}
\newacro{UWB}{ultra wide band}
\newacro{TOA}{time of arrival}
\newacro{TDOA}{time difference of arrival}
\newacro{AOA}{angle of arrival}
\newacro{RSS}{received signal strength}
\newacro{GIS}{geographical information systems}
\newacro{PF}{particle filter}
\newacro{BFI}{Bayesian Fisher information}
\newacro{BFIM}{Bayesian Fisher information matrix}
\newacro{CRB}{Cramer Rao bound}     
\newacro{BCRB}{Bayesian Cramer Rao bound}
\newacro{EZZB}{extended Ziv Zikai bound}
\newacro{WWB}{Weiss Weinstein bound}

\newacro{MSE}{mean square error}
\newacro{BMSE}{Bayesian mean square error}
\newacro{MMSE}{minimum mean square error}
\newacro{MAP}{maximum a posteriori}
\newacro{ML}{maximum likelihood}
\newacro{SNR}{signal-to-noise ratio}
\newacro{FLOP}{floating point operation}
\newacro{NLOS}{non-line-of-sight}
\newacro{HMM}{hidden Markov model}
\newacro{KF}{Kalman filter}
\newacro{PCRB}{posterior Cramer Rao bound}
\newacro{PFI}{posterior Fisher information}
\newacro{PFIM}{posterior Fisher information matrix}
\newacro{SLAM}{simultaneous localization and mapping}
\newacro{LS}{least squares}
\newacro{RHS}{right hand side}

\newacro{POCS}{projection onto convex sets}
\newacro{MDS}{multidimensional scaling}

\newacro{RMSE}{root median square error}
\newacro{LOS}{line of sight}
\newacro{WED}{wall extra delay}
\newacro{WAF}{wall attenuation factor}
\newacro{MWM}{multi-wall model}
\newacro{MNC}{minimal non-overlapping covering}
\newacro{DNC}{dense non-overlapping covering}
\newacro{RRS}{reproducible research standard}
\newacro{DRD-MMSE}{distance-reduced domain \emph{MMSE}}
\newacro{DRD-MAP}{distance-reduced domain \emph{MAP}}
\newacro{PRD-MMSE}{probability-reduced domain \emph{MMSE}}
\newacro{PRD-MAP}{probability-reduced domain \emph{MAP}}
\newacro{CDE}{cluster detection error}
\newacro{CGDE}{cluster group detection error}

\newacro{INS}{inertial navigation system}
\newacro{IMU}{inertial measurement unit}
\newacro{MEMS}{micro electro-mechanical systems}
\newacro{PDR}{pedestrian dead-reckoning}
\newacro{EKF}{extended Kalman filter}
\newacro{RBPF}{Rao-Blackwellized particle filter}
\newacro{SEM}{sensor error model}
\newacro{AHRS}{attitude and heading reference system}
\newacro{AWGN}{additive white Gaussian noise}
\newacro{AGN}{additive Gaussian noise}
\newacro{ZUPT}{zero velocity update}
\newacro{SIR}{sequential importance resampling}
\newacro{ARW}{acceleration random walk}
\newacro{BI}{bias instability}
\newacro{RRW}{rate random walk}
\newacro{PSD}{power spectrum density}

\newacro{SFS}{soft foot still}

\newacro{TTD}{total travelled distance}
\newacro{CAE}{checkpoint accumulated error}
\newacro{CIDE}{checkpoint inter-distance error}

\newacro{AFTT}{average filtering time for timestep}

\newacro{TE}{turbo equalization}

\newacro{muINS}{map-unaware INS}
\newacro{tINS}{turbo INS}
